\documentclass{article}

\usepackage{arxiv}

\usepackage[utf8]{inputenc} % allow utf-8 input
\usepackage[T1]{fontenc}    % use 8-bit T1 fonts
\usepackage{hyperref}       % hyperlinks
\usepackage{url}            % simple URL typesetting
\usepackage{booktabs}       % professional-quality tables
\usepackage{amsfonts}       % blackboard math symbols
\usepackage{nicefrac}       % compact symbols for 1/2, etc.
\usepackage{microtype}      % microtypography
\usepackage{lipsum}			% Can be removed after putting your text content
\usepackage{graphicx}
\usepackage{todonotes}
\usepackage{bm}
\usepackage{tikz}
\usepackage{amsmath}

\title{On Feature Relevance Uncertainty: A Monte Carlo Dropout Sampling Approach}

\date{}

\author{Kai Fischer \\ \texttt{fishrkai@gmail.com} \\
        \And
        Jonas Schneider \\ EFS \\ \texttt{jonas.schneider@efs-auto.com} \\
}

\makeatletter
\def\blfootnote{\gdef\@thefnmark{}\@footnotetext}
\makeatother

\begin{document}
    % Parameters
    % Imagesize

% Results 
\newcommand\imagesize{0.8\textwidth}    

% Appendix
\newcommand\imgsizeapp{0.8\textwidth}

    \maketitle
    \begin{abstract}
        Understanding decisions made by neural networks is key for the deployment of intelligent systems in real world applications. However, the opaque decision making process of these systems is a disadvantage where interpretability is essential. Many feature-based explanation techniques have been introduced over the last few years in the field of machine learning to better understand decisions made by neural networks and have become an important component to verify their reasoning capabilities. However, existing methods do not allow statements to be made about the uncertainty regarding a feature's relevance for the prediction. In this paper, we introduce Monte Carlo Relevance Propagation (MCRP) for feature relevance uncertainty estimation. A simple but powerful method based on Monte Carlo estimation of the feature relevance distribution to compute feature relevance uncertainty scores that allow a deeper understanding of a neural network's perception and reasoning.

    \end{abstract}

    \blfootnote{This work is published under the Creative Commons license CC BY-NC-4.0}
    
    %%%%%%%%%%%%%%%%%%%%%%
\section{Introduction} \label{sec:introduction}
%%%%%%%%%%%%%%%%%%%%%%

Understanding the reasoning capabilities of machine learning models on a deeper level is necessary in order to detect possible flaws that are oftentimes difficult to identify. Explainability of decisions made by neural networks in the field of computer vision is key for the deployment and safety of intelligent computer vision systems. However, due to the large number of parameters and their highly nonlinear structure, the resulting decision making process of neural networks is no longer comprehensible. To address the problem of limited transparency and to detect possible flaws of a model at an early stage, in recent years numerous methods have been proposed that are based on assigning prediction-related relevance scores to input features (i.e. pixels) creating a heatmap that allows for a better understanding of why the network made a certain decision.

Even though these methods allow a better assessment of the maturity level of machine learning based systems, they do not allow statements about how certain a network is in what it perceives. This means that they do not assign uncertainty scores to individual input features regarding their relevance for the network's decision. However, the ability to indicate uncertainty regarding the importance of input features should be a core component of machine learning based perception systems that are used in safety-critical areas to make better statements about their reliability. Until now, uncertainty estimates have only been used for network predictions as for example in the work of Gal et al. (\cite{gal2016dropout}). 

This work combines feature-based explanation with uncertainty estimation and introduces a straight forward, easy to implement, and fully unsupervised method to compute feature-wise relevance uncertainty scores based on Monte Carlo dropout sampling. In contrast to other methods that determine the features' importance for the classification using point estimates, our method allows to compute approximate feature relevance distributions to estimate mean relevance scores and relevance uncertainty scores of input features. The work of Bach et al. (\cite{bach2015lrp}) on Layer-wise Relevance Propagation (LRP) and Gal et al. (\cite{gal2016dropout}) on predictive uncertainty estimation form the theoretical basis of this paper.

In this work, we make use of LRP to assign prediction-related relevance uncertainty scores to input features. However, our sampling-based estimation of feature relevance uncertainty scores is not constrained to LRP and can increase the expressive power of many other feature-based explanation techniques. In principle, any method that exploits the network's graph structure to emphasize the importance of certain input features for the prediction can be used. Methods that can be extended relatively easily by a feature relevance uncertainty measure using a sampling approach are BagNet (\cite{brendel2019approximating}), DeepLIFT (\cite{shrikumar2017learning}), Grad-cam (\cite{selvaraju2017grad}), Guided Backpropagation (\cite{zeiler2014visualizing}), PatternNet (\cite{kindermans2017learning}), Sensitivity Analysis (\cite{baehrens2010explain}), Softmax Gradient LRP (\cite{iwana2019explaining}), and attention networks such as MAC (\cite{hudson2018compositional}), to name just a few.

In this paper, we show empirical evidence that our method allows to estimate relevance uncertainty scores for input features. To make the results easier to understand, we apply MCRP to an image classification task. However, the application of MCRP is not limited to images and can be extended to any input data.
    %%%%%%%%%%%%%%%%%%%%%%
\section{Related Work} \label{sec:related_work}
%%%%%%%%%%%%%%%%%%%%%%

MCRP combines Layer-wise Relevance Propagation (LRP) and predictive uncertainty estimation using Monte Carlo dropout sampling for the computation of feature relevance uncertainty scores. Both methods will be presented here very briefly. For a more in-depth discussion of both topics, we refer to the literature.

%%%%%%%%%%%%%%%%%%%%%%%%%%%%%%%%%%%%%%%%%%%%%%%%%%%%%%%%%%%%%%%%%%%%%%%%%%%
\subsection{Layer-wise Relevance Propagation and Deep Taylor Decomposition}
%%%%%%%%%%%%%%%%%%%%%%%%%%%%%%%%%%%%%%%%%%%%%%%%%%%%%%%%%%%%%%%%%%%%%%%%%%%

Feature-based explanation techniques such as LRP (\cite{bach2015lrp}) try to identify input features that are most relevant for a network's classification decision. Some of these methods use the network's graph structure to compute feature-wise relevance scores by decomposing the network's prediction in a layer-wise fashion from the output back to the input layer. Figure \ref{fig:lrp_network} illustrates the principle of LRP.

\usetikzlibrary{chains, positioning, decorations.pathreplacing}

% Example images
\def\groundtruth{"./assets/images/methods/example"}
\def\predmean{"./assets/images/methods/example_avg_with_gt"}

% Network architecture
\def\layersep{2cm}
\def\hsep{1cm}
\def\ilsize{3}
\def\hlsize{4}
\def\olsize{3}
\def\rootlrp{6}
\def\neuronsize{6mm}

\tikzset{>=latex}

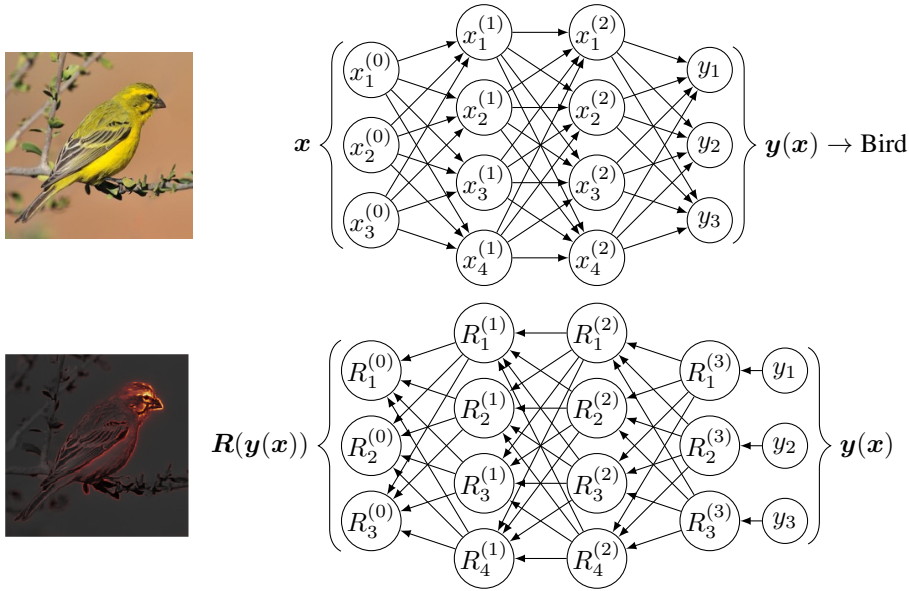
\begin{figure}[!htb]
\centering
%-------------%
% Feedforward %
%-------------%
\begin{tikzpicture}[shorten >=0pt, ->, draw=black!100, node distance=\layersep]
\tikzstyle{every pin edge}=[<-,shorten <=1pt]
\tikzstyle{neuron}=[circle, draw, fill=black!100, minimum size=\neuronsize,inner sep=0pt]
\tikzstyle{input neuron}=[neuron, fill=black!0]
\tikzstyle{hidden neuron}=[neuron, fill=black!0]
\tikzstyle{output neuron}=[neuron, fill=black!0]
\tikzstyle{annot} = [text width=4em, text centered, node distance=5mm]

\pgfmathsetmacro{\iyshift}{0.5*\ilsize-0.5*\hlsize}
\pgfmathsetmacro{\oyshift}{0.5*\olsize-0.5*\hlsize}
%%%%%%%%%%%%
% DRAW NODES
%%%%%%%%%%%%
% Draw the input layer nodes
\foreach \name / \y in {1,...,\ilsize}
    \node[input neuron] (In-\name) at (0.0cm+\hsep,-\y cm+\iyshift cm) {$x_{\y}^{(0)}$};
% Draw the hidden layer nodes
\foreach \name / \y in {1,...,\hlsize}
    \node[hidden neuron] (H0-\name) at (1.5cm+\hsep,-\y cm) {$x_{\y}^{(1)}$};
% Draw the hidden layer nodes
\foreach \name / \y in {1,...,\hlsize}
    \node[hidden neuron] (H1-\name) at (3.0cm+\hsep,-\y cm) {$x_{\y}^{(2)}$};
% Draw the output layer nodes
\foreach \name / \y in {1,...,\olsize}
    \node[hidden neuron] (Out-\name) at (4.5cm+\hsep,-\y cm+\oyshift cm) {$y_{\y}$};

%%%%%%%%%%%%%%%%%%
% DRAW CONNECTIONS
%%%%%%%%%%%%%%%%%%
% Connect every node in the input layer with every node in the hidden layer.
\foreach \source in {1,...,\ilsize}
    \foreach \dest in {1,...,\hlsize}
        \path (In-\source) edge (H0-\dest);
% Connect first with second hidden layer
\foreach \source in {1,...,\hlsize}
    \foreach \dest in {1,...,\hlsize}
        \path (H0-\source) edge (H1-\dest);
% Connect every node from the last hidden layer with the output layer
\foreach \source in {1,...,\hlsize}
    \foreach \dest in {1,...,\olsize}
        \path (H1-\source) edge (Out-\dest);

%%%%%%%%%%%%%%%%%%%%
% Annotate Network %
%%%%%%%%%%%%%%%%%%%%
\draw[-, decoration={brace,raise=0pt, amplitude=3mm}, decorate, xshift=0mm, yshift=0mm]
(Out-1.north -| Out-1.east) -- node[right=3mm] {$\bm{y}(\bm{x}) \rightarrow \textrm{Bird}$} (Out-\olsize.south -| Out-\olsize.east);
\draw[-, decoration={brace, raise=0pt, amplitude=3mm, mirror}, decorate, xshift=0mm, yshift=0mm] 
(In-1.north -| In-1.west) -- node[left=3mm] {$\bm{x}$}   (In-\ilsize.south -| In-\ilsize.west);

%% Add input image
\node[inner sep=0pt, left = 2cm of In-2] (image) {\includegraphics[width=.15\textwidth]{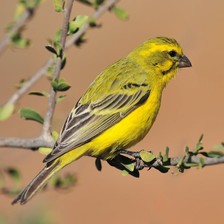}};

\begin{scope}[yshift=-40mm]
%---------------------------%
% Relevance Backpropagation %
%---------------------------%
\tikzstyle{every pin edge}=[<-,shorten <=1pt]
\tikzstyle{neuron}=[circle,draw,fill=black!100,minimum size=\neuronsize,inner sep=0pt]
\tikzstyle{input neuron}=[neuron, fill=black!0]
\tikzstyle{hidden neuron}=[neuron, fill=black!0]
\tikzstyle{output neuron}=[neuron, fill=black!0]
\tikzstyle{annot} = [text width=4em, text centered, node distance=5mm]

\pgfmathsetmacro{\iyshift}{0.5*\ilsize-0.5*\hlsize}
\pgfmathsetmacro{\oyshift}{0.5*\olsize-0.5*\hlsize}
%%%%%%%%%%%%
% DRAW NODES
%%%%%%%%%%%%
% Draw the input layer nodes
\foreach \name / \y in {1,...,\ilsize}
    \node[input neuron] (In-\name) at (0.0cm+\hsep,-\y cm+\iyshift cm) {$R_{\y}^{(0)}$};
% Draw the hidden layer nodes
\foreach \name / \y in {1,...,\hlsize}
    \node[hidden neuron] (H0-\name) at (1.5cm+\hsep,-\y cm) {$R_{\y}^{(1)}$};
% Draw the hidden layer nodes
\foreach \name / \y in {1,...,\hlsize}
    \node[hidden neuron] (H1-\name) at (3.0cm+\hsep,-\y cm) {$R_{\y}^{(2)}$};
%% Draw the output layer nodes
%\foreach \name / \y in {1,...,\olsize}
%    \node[hidden neuron] (Out-\name) at (4.5cm+\hsep,-\y cm+\oyshift cm) {$y_{\y}$};

%%%%%%%%%%%%%
% Draw the output layer nodes
\foreach \name / \y in {1,...,\olsize}
    \node[hidden neuron] (Out-\name) at (4.5cm+\hsep,-\y cm+\oyshift cm) {$R_{\y}^{(3)}$};
% Draw the output layer nodes
\foreach \name / \y in {1,...,\olsize}
    \node[hidden neuron] (Out2-\name) at (5.5cm+\hsep,-\y cm+\oyshift cm) {$y_{\y}$};
%%%%%%%%%%%%%

%%%%%%%%%%%%%%%%%%
% DRAW CONNECTIONS
%%%%%%%%%%%%%%%%%%
% Connect every node in the input layer with every node in the hidden layer.
\foreach \source in {1,...,\ilsize}
    \foreach \dest in {1,...,\hlsize}
        \path[<-] (In-\source) edge (H0-\dest);
% Connect first with second hidden layer
\foreach \source in {1,...,\hlsize}
    \foreach \dest in {1,...,\hlsize}
        \path[<-] (H0-\source) edge (H1-\dest);
% Connect every node from the last hidden layer with the output layer
\foreach \source in {1,...,\hlsize}
    \foreach \dest in {1,...,\olsize}
        \path[<-] (H1-\source) edge (Out-\dest);

% Connect every node from the last hidden layer with the output layer
\foreach \i in {1,...,\olsize}
    \path[<-] (Out-\i) edge (Out2-\i);

%%%%%%%%%%%%%%%%%%%%
% Annotate Network %
%%%%%%%%%%%%%%%%%%%%
\draw[-, decoration={brace,raise=0pt, amplitude=3mm}, decorate, xshift=0mm, yshift=0mm]
(Out2-1.north -| Out2-1.east) -- node[right=3mm] {$\bm{y}(\bm{x})$} (Out2-3.south -| Out2-3.east);
\draw[-, decoration={brace, raise=0pt, amplitude=3mm, mirror}, decorate, xshift=0mm, yshift=0mm] 
(In-1.north -| In-1.west) -- node[left=3mm] {$\bm{R}(\bm{y}(\bm{x}))$}   (In-3.south -| In-3.west);

%% Add input image
\node[inner sep=0pt, left = 2cm of In-2] (image) {\includegraphics[width=.15\textwidth]{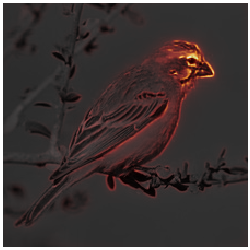}};

\end{scope}
\end{tikzpicture}
\caption{Illustration of the unsupervised relevance propagation process that consists of a feedforward pass and a subsequent backward relevance distribution pass. As can be seen here in the picture, pixels that belong to the bird's head area have been assigned great relevance. For reference, the original image has been converted to grayscale and placed behind the feature relevance scores. Figure loosely adapted from \cite{montavon2017dtd}.}
\label{fig:lrp_network}
\end{figure}

Deep Taylor Decomposition (DTD, \cite{montavon2017dtd}) builds on LRP and provides a mathematically well-founded method for assigning relevance scores to input features. DTD explains classification decisions relative to the state of maximum uncertainty which means that it identifies input features which are pivotal for a certain prediction (\cite{samek2017explainable}). DTD can be used to derive various relevance propagation rules for the decomposition of network predictions, including the $z^+$-rule. The $z^+$-rule redistributes the relevance scores as follows

\begin{equation}\label{eq:z_plus_rule}
    R_i^{(l)} = \sum_{j} \frac{x_i^{(l)} w_{ij}^+}{\sum_{i'} x_{i'}^{(l)} w_{i'j}^+} R_j^{(l+1)}
\end{equation}

Here $R_i^{(l)}$ stands for the relevance score associated with neuron $i$ in layer $l$. The relevance scores of the last layer, where $l=L$, are equal to the network predictions and therefore $\bm{R}^{(L)}=\bm{y}(\bm{x})$ applies. The relevance scores of the first layer, where $l=0$, represent the pixel's importance to the classification decision. The expression $w_{ij}^+$ represents the positive weights\footnote{The $z^+$-rule ignores negative weights during the relevance propagation by setting them to zero.} connecting neurons $i$ and $j$ of the adjacent layers $l$ and $l+1$. The activation of neuron $i$ in layer $l$ is denoted by $x_i^{(l)}$. From Equation \ref{eq:z_plus_rule} the following property can be derived:

\begin{equation}\label{eq:conservation}
    \sum_p R_p^{(0)} = \dots = \sum_i R_i^{(l)} = \sum_j R_j^{(l+1)} = \dots = \sum_k R_k^{(L)}
\end{equation}

From Equation \ref{eq:conservation} it follows that relevance scores are also assigned to hidden layer feature representations. This is especially interesting for feature maps of convolutional neural networks.

As was already the case for the network's prediction, the assignment of feature-wise relevance scores are point estimates with limited explanatory power compared to a pixel-wise relevance distribution. This can be seen as a disadvantage if more detailed information about the network's capability is desired since no statements can be made about how confident the network is regarding the importance of certain input features for the classification decision.

%%%%%%%%%%%%%%%%%%%%%%%%%%%%%%%%%%%%%%%%%%%%%%%%%
\subsection{Estimation of predictive uncertainty} \label{sec:predictive_uncertainty}
%%%%%%%%%%%%%%%%%%%%%%%%%%%%%%%%%%%%%%%%%%%%%%%%%

Dropout as a Bayesian approximation for representing model uncertainty is a widely used and in many areas successfully applied method for estimating the predictive uncertainty of deep neural networks (\cite{gal2016dropout}). Gal et al. showed that almost any network trained with some sort of stochastic regularization technique such as dropout allows to estimate the predictive mean $\mathbb{E}[\bm{y}]$ and predictive variance $Var[\bm{y}]$ for a given input $\bm{x}$. 

Estimates for the mean and variance can be obtained by performing multiple stochastic forward passes, which means that dropout remains activated during inference. In detail this means that the inference is repeated for $T$ iterations to generate independent and identically distributed empirical samples $\{\hat{\bm{y}}_1(\bm{x}), \dots, \hat{\bm{y}}_T(\bm{x})\}$ from an approximate predictive distribution (\cite{gal2016uncertainty}). An empirical estimator for the predictive mean and the predictive variance of this approximate predictive distribution is given by:

\begin{equation}\label{eq:predictive_mean}
\mathbb{E}[\bm{y}] \approx \frac{1}{T}\sum_{t=1}^T\hat{\bm{y}}_t(\bm{x})
\end{equation}

\begin{equation}\label{eq:predictive_variance}
Var[\bm{y}] \approx \frac{1}{T}\sum_{t=1}^T\hat{\bm{y}}_t(\bm{x})^2 - \mathbb{E}[\bm{y}]^2
\end{equation}

The sample mean in Equation \ref{eq:predictive_mean} can be interpreted as model averaging (\cite{srivastava2014dropout}). Equation \ref{eq:predictive_variance} corresponds to the sample variance that reflects the model's predictive uncertainty associated with a decision. Even though this method provides a measure of uncertainty for the network's prediction it does not allow statements to be made about how confident the network is in what it perceives on a lower level.

    %%%%%%%%%%%%%%%%%%%%%%%%%%%%%%%%%%%%%%%%%%%
\section{Monte Carlo Relevance Propagation} \label{sec:methods} % methods section
%%%%%%%%%%%%%%%%%%%%%%%%%%%%%%%%%%%%%%%%%%%

In this section we show how to combine Layer-wise Relevance Propagation (LRP, \cite{bach2015lrp}) and predictive uncertainty estimation (\cite{gal2016dropout}) to compute approximate pixel-wise relevance distributions that allow a more detailed understanding of the network's confidence regarding the importance of input features for the classification.

For the estimation of relevance uncertainty scores, we integrate Monte Carlo dropout sampling into LRP by treating the computation of feature relevance scores as the result of a segmentation task. The segmentation network (compare Figure \ref{fig:lrp_network_v2}) associated with this task hereby distinguishes between relevant and irrelevant input features. This assumption allows that the estimation of feature-wise uncertainty scores can be performed in the same way as described in Section \ref{sec:predictive_uncertainty}.

\usetikzlibrary{chains, positioning, decorations.pathreplacing}

% Example images
\def\groundtruth{"./assets/images/methods/example"}
\def\predmean{"./assets/images/methods/example_avg_with_gt"}

% Network architecture
\def\layersep{2cm}
\def\hsep{1cm}
\def\ilsize{3}
\def\hlsize{4}
\def\olsize{3}
\def\rootlrp{6}
\def\neuronsize{6mm}
\newcommand{\dropoutrate}{0.3}

\tikzset{>=latex}

\begin{figure}[!htb]
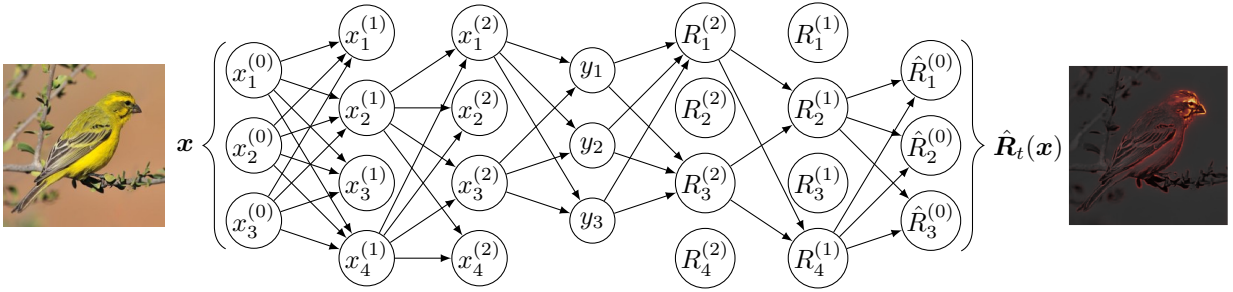

\centering
\begin{tikzpicture}[shorten >=0pt, ->, draw=black!100, node distance=\layersep]
\tikzstyle{every pin edge}=[<-,shorten <=1pt]
\tikzstyle{neuron}=[circle, draw, fill=black!100, minimum size=\neuronsize, inner sep=0pt]
\tikzstyle{input neuron}=[neuron, fill=black!0]
\tikzstyle{hidden neuron}=[neuron, fill=black!0]
\tikzstyle{output neuron}=[neuron, fill=black!0]
\tikzstyle{annot} = [text width=4em, text centered, node distance=5mm]    

\pgfmathsetmacro{\iyshift}{0.5*\ilsize-0.5*\hlsize}
\pgfmathsetmacro{\oyshift}{0.5*\olsize-0.5*\hlsize}

%%%%%%%%%%%%
% DRAW NODES
%%%%%%%%%%%%
% Draw the input layer nodes
\foreach \name / \y in {1,...,\ilsize}
    \node[input neuron] (In-\name) at (0.0cm+\hsep,-\y cm+\iyshift cm) {$x_{\y}^{(0)}$};
% Draw the hidden layer nodes
\foreach \name / \y in {1,...,\hlsize}
    \node[hidden neuron] (H0-\name) at (1.5cm+\hsep,-\y cm) {$x_{\y}^{(1)}$};
% Draw the hidden layer nodes
\foreach \name / \y in {1,...,\hlsize}
    \node[hidden neuron] (H1-\name) at (3.0cm+\hsep,-\y cm) {$x_{\y}^{(2)}$};
% Draw the output layer nodes
\foreach \name / \y in {1,...,\olsize}
    \node[hidden neuron] (Out-\name) at (4.5cm+\hsep,-\y cm+\oyshift cm) {$y_{\y}$};
% Draw nodes from hidden layer 2
\foreach \name / \y in {1,...,\hlsize}
    \node[hidden neuron] (H11-\name) at (6.0cm+\hsep,-\y cm) {$R_{\y}^{(2)}$};
% Draw nodes from hidden layer 1
\foreach \name / \y in {1,...,\hlsize}
    \node[hidden neuron] (H01-\name) at (7.5cm+\hsep,-\y cm) {$R_{\y}^{(1)}$};
% Draw nodes from input layer
\foreach \name / \y in {1,...,\ilsize}
    \node[hidden neuron] (In1-\name) at (9.0cm+\hsep,-\y cm+\iyshift cm) {$\hat{R}_{\y}^{(0)}$};

%%%%%%%%%%%%%%%%%%
% DRAW CONNECTIONS
%%%%%%%%%%%%%%%%%%

\def\arrB{{0,0,1,0,1}}
\def\arrC{{0,1,0,1,0}}

% Input to hidden
\foreach \i in {1,...,\ilsize}
{
    \foreach \j in {1,...,\hlsize}
    {
            \path (In-\i) edge (H0-\j);
    }
}

% Hidden to hidden
\foreach \i in {1,...,\hlsize}
{
    \pgfmathsetmacro\draw{ifthenelse(\arrB[\i]==1, "1", "0")}
    \ifnum\draw=1
    \foreach \j in {1,...,\hlsize}
    {
            \path (H0-\i) edge (H1-\j);
    }
    \fi
}

% Hidden to output
\foreach \i in {1,...,\hlsize}
{
    \pgfmathsetmacro\draw{ifthenelse(\arrC[\i]==1, "1", "0")}
    \ifnum\draw=1
    \foreach \j in {1,...,\olsize}
    {
            \path (H1-\i) edge (Out-\j);
    }
    \fi
}

% Output to hidden
\foreach \i in {1,...,\olsize}
{
    \foreach \j in {1,...,\hlsize}
    {
        \pgfmathsetmacro\draw{ifthenelse(\arrC[\j]==1, "1", "0")}
        \ifnum\draw=1
            \path (Out-\i) edge (H11-\j);
        \fi
    }
}

% Hidden to hidden
\foreach \i in {1,...,\hlsize}
{
    \pgfmathsetmacro\draw{ifthenelse(\arrC[\i]==1, "1", "0")}
    \ifnum\draw=1
    \foreach \j in {1,...,\hlsize}
    {
        \pgfmathsetmacro\draw{ifthenelse(\arrB[\j]==1, "1", "0")}
        \ifnum\draw=1
            \path (H11-\i) edge (H01-\j);
        \fi
    }
    \fi
}

% Hidden to input
\foreach \i in {1,...,\hlsize}
{
    \pgfmathsetmacro\draw{ifthenelse(\arrB[\i]==1, "1", "0")}
    \ifnum\draw=1
    \foreach \j in {1,...,\ilsize}
    {
        \path (H01-\i) edge (In1-\j);
    }
    \fi
}

%%%%%%%%%%%%%%%%%%%%
% Annotate Network %
%%%%%%%%%%%%%%%%%%%%
\draw[-, decoration={brace,raise=0pt, amplitude=3mm}, decorate, xshift=0mm, yshift=0mm]
(In1-1.north -| In1-1.east) -- node[right=3mm] {$\hat{\bm{R}}_t(\bm{x})$} (In1-3.south -| In1-3.east);
\draw[-, decoration={brace, raise=0pt, amplitude=3mm, mirror}, decorate, xshift=0mm, yshift=0mm] 
(In-1.north -| In-1.west) -- node[left=3mm] {$\bm{x}$}   (In-3.south -| In-3.west);

% Input image
\node[inner sep=0pt, left = 0.8cm of In-2] (image) {\includegraphics[width=.13\textwidth]{\groundtruth}};

% Output feature relevance prediction
\node[inner sep=0pt, right = 1.4cm of In1-2] (image) {\includegraphics[width=.13\textwidth]{\predmean}};

\end{tikzpicture}
\caption{Drawing of the feature relevance prediction network. The information processing consists of two parts. A classical feedforward pass to arrive at predictions $y_i$ and the subsequent processing of the predictions to compute the estimates of feature-wise relevance scores $\hat{R}_i^{(0)}$ at the output layer. The drawing shows that dropped neurons do not contribute to deeper-layer neurons and therefore neither receive relevance nor can they pass on relevance to subsequent neurons. Please note that the same dropout mask is used for both network parts. Again, the original image has been converted to grayscale and is placed behind the computed relevance scores for reference.}
\label{fig:lrp_network_v2}
\end{figure}

The information processing of this network consists of a feedforward pass to arrive at predictions $y_i$, which are then processed further to compute the estimation of feature-wise relevance scores $\hat{R}_i^{(0)}$ at the output layer. This process is fully unsupervised. Note that the first network part shares all weights and activations with the second part of the network and that only the instructions regarding the neuron-wise transformation differ. While in the first half of the network, the neuron-wise transformations are given by 

\begin{equation}\label{eq:relu_activation}
    x_{j,t}^{(l+1)} = \max(0, \sum_i x_i^{(l)} \tilde{w}_{ij} + b_j)
\end{equation}

the transformations of the second part are given by 

\begin{equation}\label{eq:z_plus_rule_v2}
    \hat{R}_{i,t}^{(l)} = \sum_{j} \frac{x_i^{(l)} \tilde{w}_{ij,t}^+}{\sum_{i'} x_{i'}^{(l)} \tilde{w}_{i'j,t}^+} \hat{R}_{j,t}^{(l+1)}
\end{equation}

Here, $t$ indicates the current sampling iteration. The randomly drawn parameter configuration of the current subnetwork is denoted by $\tilde{\bm{w}}$. The network's output, $\hat{\bm{R}}_t(\bm{x})$, now represents the estimated feature-wise relevance scores.

The results of $T$ stochastic forward passes are i.i.d. feature relevance scores $\{\hat{\bm{R}}_1(\bm{x}), ..., \hat{\bm{R}}_T(\bm{x})\}$. The set of feature relevance scores are empirical samples from an approximate feature relevance distribution and are used to estimate the mean feature relevance scores $\mathbb{E}[\bm{R}]$ and the feature relevance uncertainty scores $\text{Var}[\bm{R}]$ (compare Figure \ref{fig:mcrp_example}). The mean feature relevance scores of the approximate relevance score distribution is estimated by

\begin{eqnarray}\label{eq:mean_relevance_score}
    \mathbb{E}[\bm{R}] \approx \frac{1}{T} \sum_{t=1}^{T} \hat{\bm{R}}_t(\bm{x})
\end{eqnarray}

In order to obtain information about feature relevance uncertainty scores reflecting the network's state of having limited knowledge on the exact importance of a feature for the prediction, we compute the relevance uncertainty according to

\begin{equation}\label{eq:relevance_uncertainty_score}
    \text{Var}[\bm{R}] \approx \frac{1}{T} \sum_{t=1}^{T} \hat{\bm{R}}_t(\bm{x})^2 - \mathbb{E}[\bm{R}]^2
\end{equation}

In the following we denote by $\bm{\mu} = \mathbb{E}[\bm{R}]$ the mean feature relevance scores and by $\bm{\sigma} = \sqrt{\text{Var}[\bm{R}]}$ the uncertainty in units of relevance.

\newcommand{\imgsize}{2cm}
\newcommand{\nimages}{6}

\begin{figure}[!htb]
\centering
    \begin{tikzpicture}
        \foreach \X [count=\Z]in {1, ..., \nimages} {
            \ifnum\X=\nimages
                \node[label=below:$\{\hat{\bm{R}}_1{,} \ \hat{\bm{R}}_2{,} \ \dots{,} \ \hat{\bm{R}}_T\}$, opacity=0.5] at (0, 0, 0.7*\Z) 
                {\includegraphics[width=\imgsize,height=\imgsize] {./assets/images/methods/example_0\X.png}};
            \else 
                \node[opacity=0.5] at (0, 0, 0.7*\Z) 
                {\includegraphics[width=\imgsize,height=\imgsize] {./assets/images/methods/example_0\X.png}};
            \fi
        }
        \node[label=below:$\bm{\mu}$] at (3, -1.414, 0.5) 
        {\includegraphics[width=\imgsize,height=\imgsize] {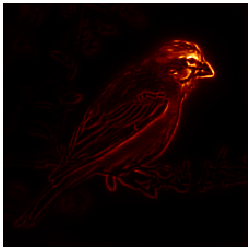}};

        \node[label=below:$\bm{\sigma}$] at (6, -1.414, 0.5) 
        {\includegraphics[width=\imgsize,height=\imgsize] {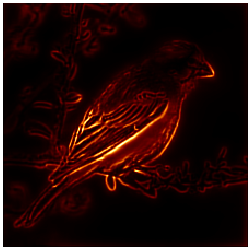}};
    \end{tikzpicture}
    \caption{Example showing a batch of feature relevance estimates $\hat{\bm{R}}_t$ on the left. The importance of every pixel with respect to the classification is represented by a distribution of relevance scores. The approximated feature relevance distributions are then used to estimate mean feature relevance scores $\bm{\mu}$ and relevance uncertainty scores $\bm{\sigma}$.}
\label{fig:mcrp_example}
\end{figure}

%%%%%%%%%%%%%%%%%%%%%%%%%%%%%
\subsection{Analysis metrics}
%%%%%%%%%%%%%%%%%%%%%%%%%%%%%

In addition to the computation of mean relevance scores and relevance uncertainty scores, the estimation of an approximate feature-wise relevance distribution for each pixel also allows the use of more advanced metrics to gain new insights regarding the network's reasoning capabilities. Metrics that combine mean relevance scores and relevance uncertainty scores are particularly interesting as they allow highlighting specific characteristics that reflect the network's understanding of a given input. 

\paragraph{Signal-to-noise ratio:} The signal-to-noise ratio, $\frac{\bm{\mu}}{\bm{\sigma}}$, particularly highlights input features that were most certainly relevant to the network's classification decision by weighting the relevance scores by the corresponding uncertainty. Signal-to-noise scores can be considered a refinement of the original results and are more precise than the classical LRP method with regard to the localization of relevant features.

\paragraph{Confusion:} We refer to $\bm{\mu} \cdot \bm{\sigma}$ as the confusion. This metric allows identifying features that are apparently of high importance for the network's classification decision, but at the same time are also subject to great uncertainty. The confusion score can be used to highlight areas in the image where the network has difficulty classifying features as relevant with high certainty and can therefore be understood as a kind of robustness metric.

%%%%%%%%%%%%%%%%%%%%%%%%
\subsection{Experiments}
%%%%%%%%%%%%%%%%%%%%%%%%

In our experiments, we performed an empirical evaluation of mean feature relevance scores, feature relevance uncertainty scores, signal-to-noise scores, and confusion scores. To emphasize the network's uncertainty with respect to the input features and to make the results more intuitive, we tested our method on images showing animals that are part of the ImageNet dataset (\cite{deng2009imagenet}). We selected four different categories of animals, namely cats, dogs, birds, and spiders. Within each of these categories, there are various subclasses, such as different breeds of cats and dogs, which the network had to learn to distinguish during the training.

For the evaluation of feature relevance uncertainty scores, we used a VGG-16 network (\cite{simonyan2014very}) that has been pre-trained on the ImageNet dataset. We implemented MCRP using Tensorflow 2.0 (\cite{tensorflow2015whitepaper}). For the computation of relevance scores, we used the $z^+$-rule (see Equation \ref{eq:z_plus_rule}) to explain individual predictions as this rule has achieved empirically good results for complex machine learning models (\cite{montavon2017dtd}). In the experiment, we used the same dropout rate of 0.5 that was used for training the VGG model. To approximate the feature relevance distributions the number of iterations $T$ was set to 100 \footnote{In general, a much smaller number of samples can be used to generate reasonably good estimates of relevance uncertainty scores.}. We would like to emphasize here that the results shown in this work are not only sensitive to the selected relevance propagation rule but also to the training procedure and the neural network's architectural details.

To compute the relevance map for a given input image, the relevance scores of each pixel’s color channels were averaged. The estimated relevance scores of each individual relevance map were then linearly normalized according to the following formula

\begin{eqnarray}
    \hat{\bm{R}}_t \leftarrow \frac{\hat{\bm{R}}_t - \text{min}(\hat{\bm{R}}_t)}{\text{max}(\hat{\bm{R}}_t) - \text{min}(\hat{\bm{R}}_t)}
\end{eqnarray}

The normalized pixel-wise relevance score maps were then used to determine mean relevance scores and relevance uncertainty scores.
       % 1) describe experiment
    %%%%%%%%%%%%%%%%%
\section{Results} \label{sec:results}
%%%%%%%%%%%%%%%%%

This section presents qualitative results obtained using the metrics introduced in Section \ref{sec:methods}. The subsequent figures show from left to right the original image, mean feature relevance, feature relevance uncertainty, signal-to-noise, and confusion scores. For reference, the original image was placed behind the computed feature scores in grayscale. Further examples can be found in Appendix \ref{sec:appendix}.

\begin{figure}[!htb]
	\centering
	\includegraphics[width=\imagesize]{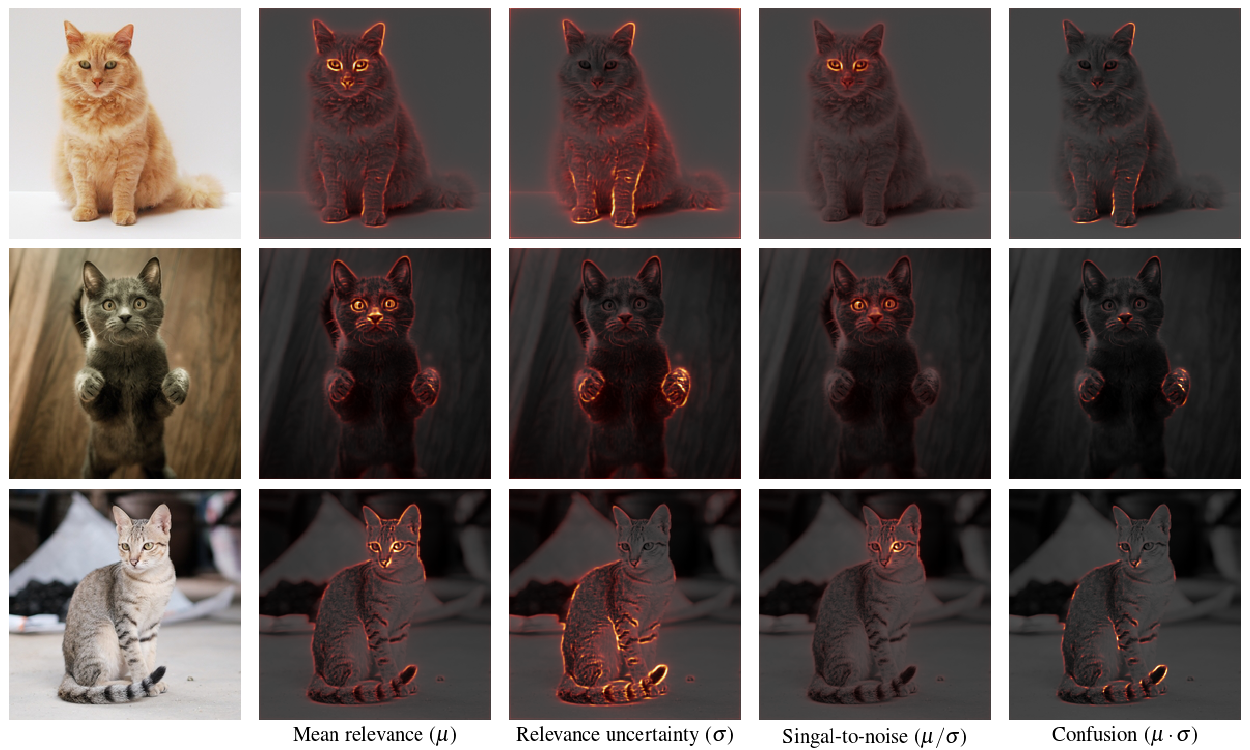}
	\caption{Results for the cat class.}
	\label{fig:cat}
\end{figure}

\begin{figure}[!htb]
	\centering
	\includegraphics[width=\imagesize]{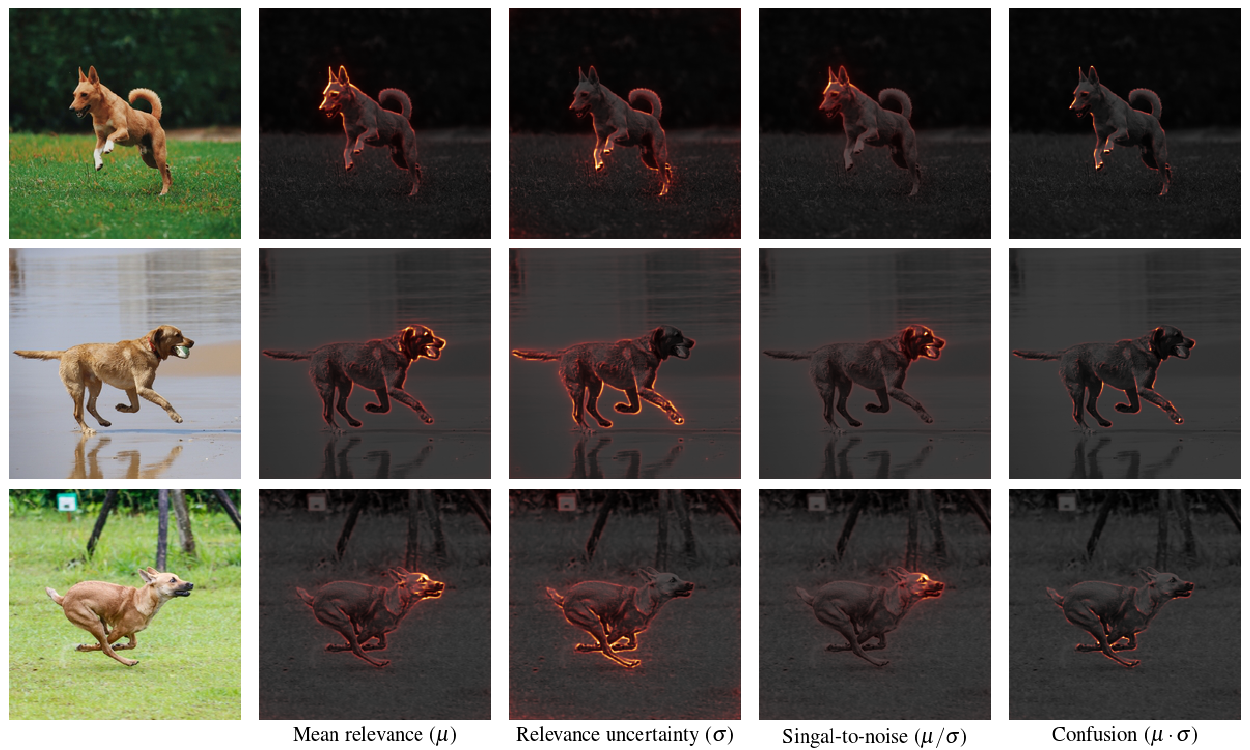}
	\caption{Results for the dog class.}
	\label{fig:dog}
\end{figure}

\begin{figure}[!htb]
	\centering
	\includegraphics[width=\imagesize]{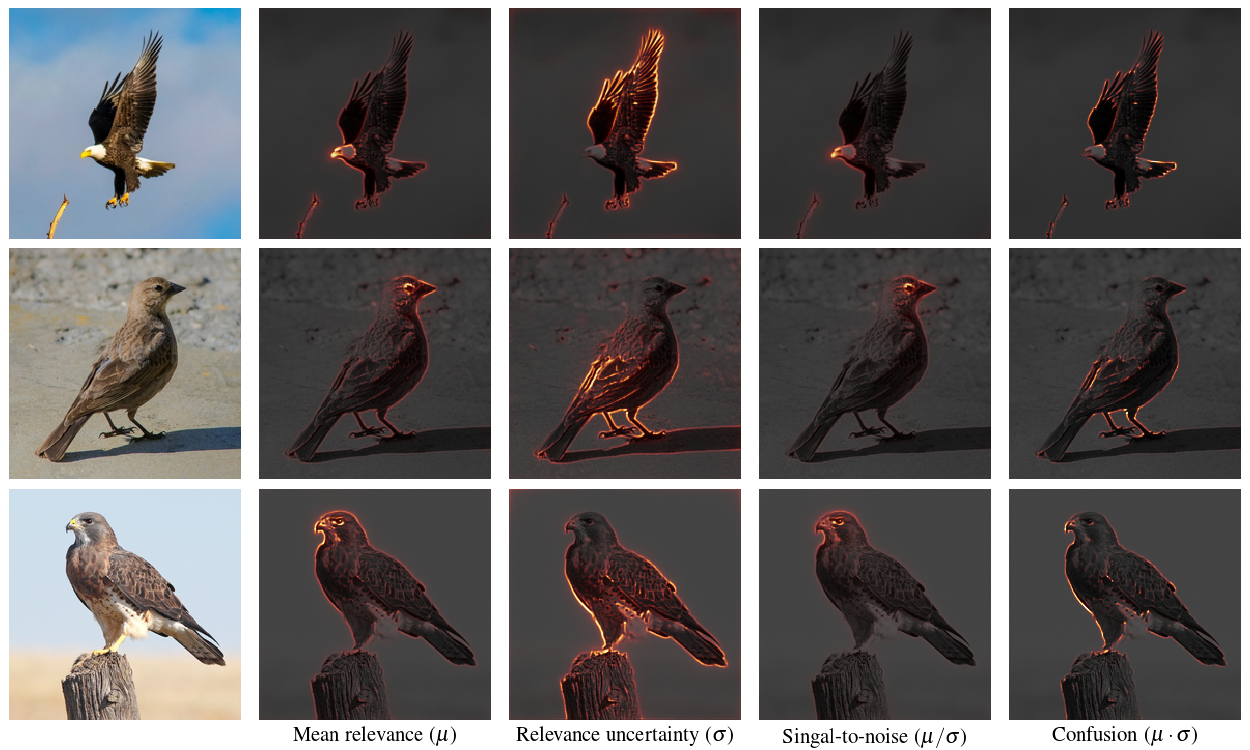}
	\caption{Results for the bird class.}
	\label{fig:bird}
\end{figure}

\begin{figure}[!htb]
	\centering
	\includegraphics[width=\imagesize]{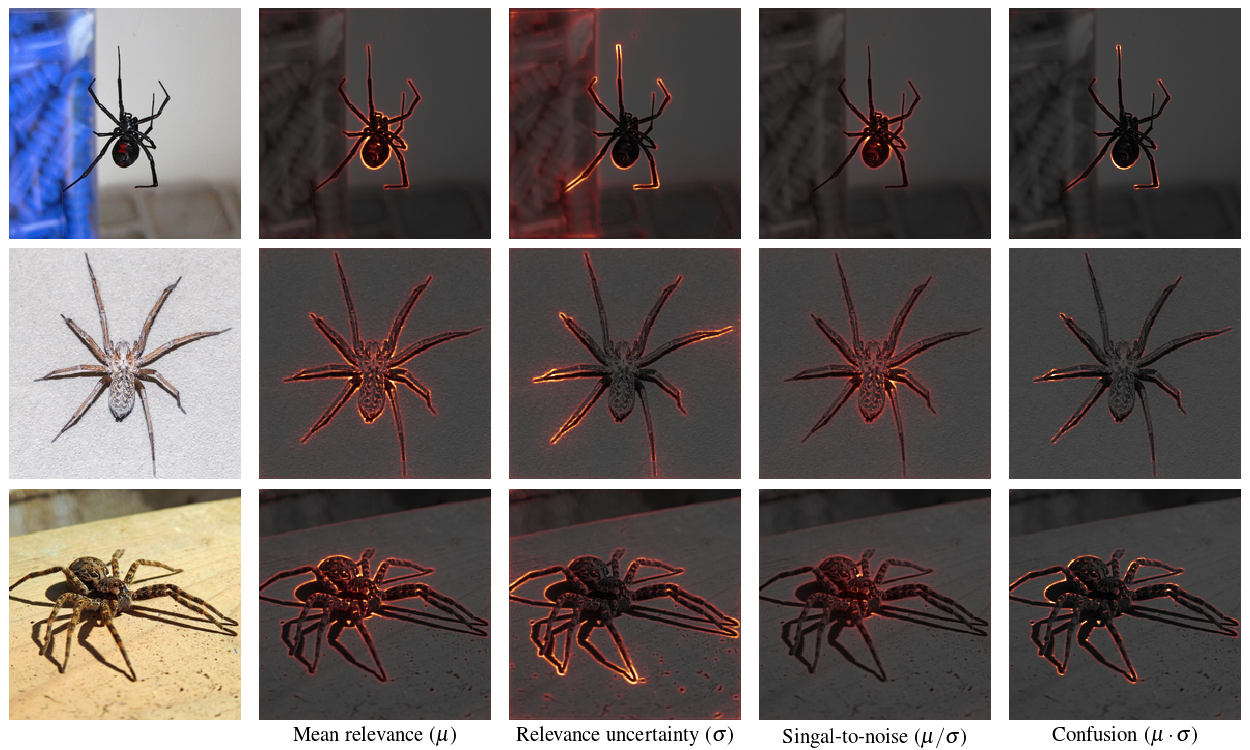}
	\caption{Results for the spider class.}
	\label{fig:spider}
\end{figure}

In Figure \ref{fig:cat}, \ref{fig:dog}, \ref{fig:bird}, and \ref{fig:spider}, it can be observed that most of the relevance is assigned to the target object. The mean feature relevance scores show that the network assigns especially high importance to the animal's head or face, indicating, that features from these areas were particularly relevant for the network's prediction. Little relevance is assigned to other regions that are not part of the object to be detected, even if these areas contain strong gradients.

Relevance uncertainty scores are also predominantly assigned to the target object. By looking at the feature relevance uncertainty scores of Figure \ref{fig:cat} and \ref{fig:dog} it turns out that for cats and dogs high uncertainty is assigned to the legs and body below the head with a few exceptions for the ears, which are also assigned relatively high uncertainty scores. Similar observations can be made for the results showing birds. Here, the plumage and legs are subject to a high degree of uncertainty. The results for images containing spiders show very similar properties. Again, the legs trigger a high degree of uncertainty in the network regarding their importance for the classification. It appears that features shared by animals of the same group are subject to a high degree of uncertainty. In contrast to mean feature relevance scores, relevance uncertainty scores are more often assigned to regions outside the object to be detected. This is especially true for areas with strong gradients.

Even more information can be extracted by the combination of mean feature relevance scores and feature relevance uncertainty scores. The results show, that the signal-to-noise ratio refines relevance scores and better identifies regions in the image that played a crucial role in the classification process. The signal-to-noise ratio allows to clearly identify particularly relevant features even if the mean feature relevance scores are widely scattered (compare \ref{fig:spider}). As can be seen from the results the signal-to-noise scores are high especially in the head and face region.

The confusion score, which can be interpreted as an indicator of perceptual robustness in image recognition tasks, highlights especially those features that were apparently important for the network's classification decision, but at the same time were also associated with a higher degree of uncertainty. The results show that in most cases the network assigns overall very few confusion scores of great magnitude to input features, and that they appear especially in areas with high gradients.
       % 2) show results of experiment 
    %%%%%%%%%%%%%%%%%%%%
\section{Discussion} 
%%%%%%%%%%%%%%%%%%%%

Our method allows computing mean feature relevance scores from estimated feature relevance distributions. Compared to other feature-based explanation techniques such as LRP (\cite{bach2015lrp}), which determines relevance scores based on a single measurement, mean feature relevance scores allow better statements about the true relevance of features.

Our results provide empirical evidence that the variance extracted from the feature relevance distribution can be considered to represent an uncertainty measure regarding the input features' importance for the prediction. This feature relevance uncertainty allows many novel evaluation methods to better assess the reasoning capabilities of neural networks. We showed empirically that signal-to-noise scores allow to further refine our results by particularly emphasizing areas relevant for the prediction. This may also be of interest in the area of feature selection, as feature relevance uncertainty scores allow relevant features to be selected with greater precision. We have also shown that confusion scores highlight features where the network has difficulty classifying them as relevant with high certainty. In our experiments, the confusion scores indicated that the network was able to cope with most of the image content and that in the majority of cases it could distinguish between important and unimportant features. At this stage of understanding, we believe that the confusion score can also be understood as a metric for perceptual robustness with many potentially interesting applications.

Our results indicate that neural networks, such as humans, can be unsure about features belonging to a certain class. This is especially noticeable when classes share characteristics that are not discriminatory enough for the network to make a decision easily. The findings show that features shared by animals of the same family are subject to a high degree of uncertainty. This makes sense, since these characteristics alone make the classification much more difficult as they can also be found in other classes. Here, parallels can be drawn between deep neural networks and human perception as humans also recognize other people mainly by their face or head and not by other body parts such as the arm or foot as this would be cognitively much more demanding if not impossible.

From the results, it can also be concluded that during the training process networks learn to associate certain unique characteristics with a class. For images containing animals, these characteristics are oftentimes the animal's head or face. This may be the case as these characteristics are the most simple and discriminating features that allow the network to distinguish between classes. The use of simple if not the simplest features to identify a class can be seen as a disadvantage since the network does not necessarily learn a global concept of what it sees during the training process. This behavior can be considered as a kind of overfitting with respect to the simplest features.

It also seems possible to use feature relevance uncertainty scores as an additional evaluation metric alongside other metrics such as loss and accuracy. During the training process, the network should become more familiar with the task and therefore progressively assign smaller uncertainty scores to input features. This information can then be used to determine the network's overall feature relevance uncertainty on the validation set during the training progress.

By determining an accumulated feature relevance uncertainty score on a test dataset it would also be possible to compare different networks in terms of their perceptual confidence level. In safety-critical areas, neural networks that identify relevant features with a high degree of certainty and thereby express that they understand fairly well what they perceive, should be prioritized.

MCRP allows computing relevance uncertainty scores not only for input features but also for their abstract hidden layer representations. In convolutional neural networks, for example, certain features maps could be examined in terms of feature representation uncertainty as well.    % 3) discuss experiments
    %%%%%%%%%%%%%%%%%%%%%%%%%%%%%%%%%%%%%%%%
\section{Conclusion and Future Research} \label{sec:conclusion}
%%%%%%%%%%%%%%%%%%%%%%%%%%%%%%%%%%%%%%%%

We showed that our method, which combines Layer-wise Relevance Propagation (\cite{bach2015lrp}) and the concept of predictive uncertainty estimation (\cite{gal2016dropout}), allows estimating feature relevance uncertainty scores. In contrast to the predictive uncertainty, which can be considered as a high-level uncertainty measure, feature relevance uncertainty scores provide low-level uncertainty information about the network.

The estimation of an approximate feature relevance distribution allows for completely new evaluation methods and provides novel insights on how networks reason about a given input. Using MCRP it is now not only possible to know when the system is uncertain, but also why. Besides knowing which features primarily contributed to the classification decision, uncertainty scores reflect the network's confidence regarding the affiliation of certain features to a class. Feature relevance uncertainty scores may also help to understand if a model is falsely over-confident by comparing the level of uncertainty with the model's prediction. There should not be a high predictive confidence and high feature relevance uncertainty scores at the same time.

Compared to other relevance propagation methods, the explanatory power of our empirically determined estimates is significantly greater as they allow for a more detailed feature-based decision analysis. Feature relevance distributions are superior as they have more expressive power and allow a deeper understanding of a model's reasoning capabilities and thus makes the behavior of neural networks more transparent.

Existing feature-based explanation techniques do not allow for feature-wise uncertainty representations as they rely on pure point estimates regarding the importance of input features. However, these methods could be extended using a sampling approach as introduced in this paper, which might also lead to many new and interesting insights.

With this work, we are just scratching the surface of many applications using MCRP
and we think that feature relevance uncertainty scores can be of great interest for many computer vision systems operating in safety critical domains such as the medical area or the field of autonomous driving. We expect the results to be even more interesting if current state-of-the-art network architectures are used. We hope to see further research in this area and refined versions of MCRP that allow for an even better understanding of what neural networks are capable of.
    % 4) derivation of general statements
    \bibliographystyle{unsrt}
    \bibliography{references}  %%% Remove comment to use the external .bib file (using bibtex).
    
    \newpage
    \appendix
    \section{Additional results figures}\label{sec:appendix}

\begin{figure}[!htb]
	\centering
    \includegraphics[width=\imagesize]{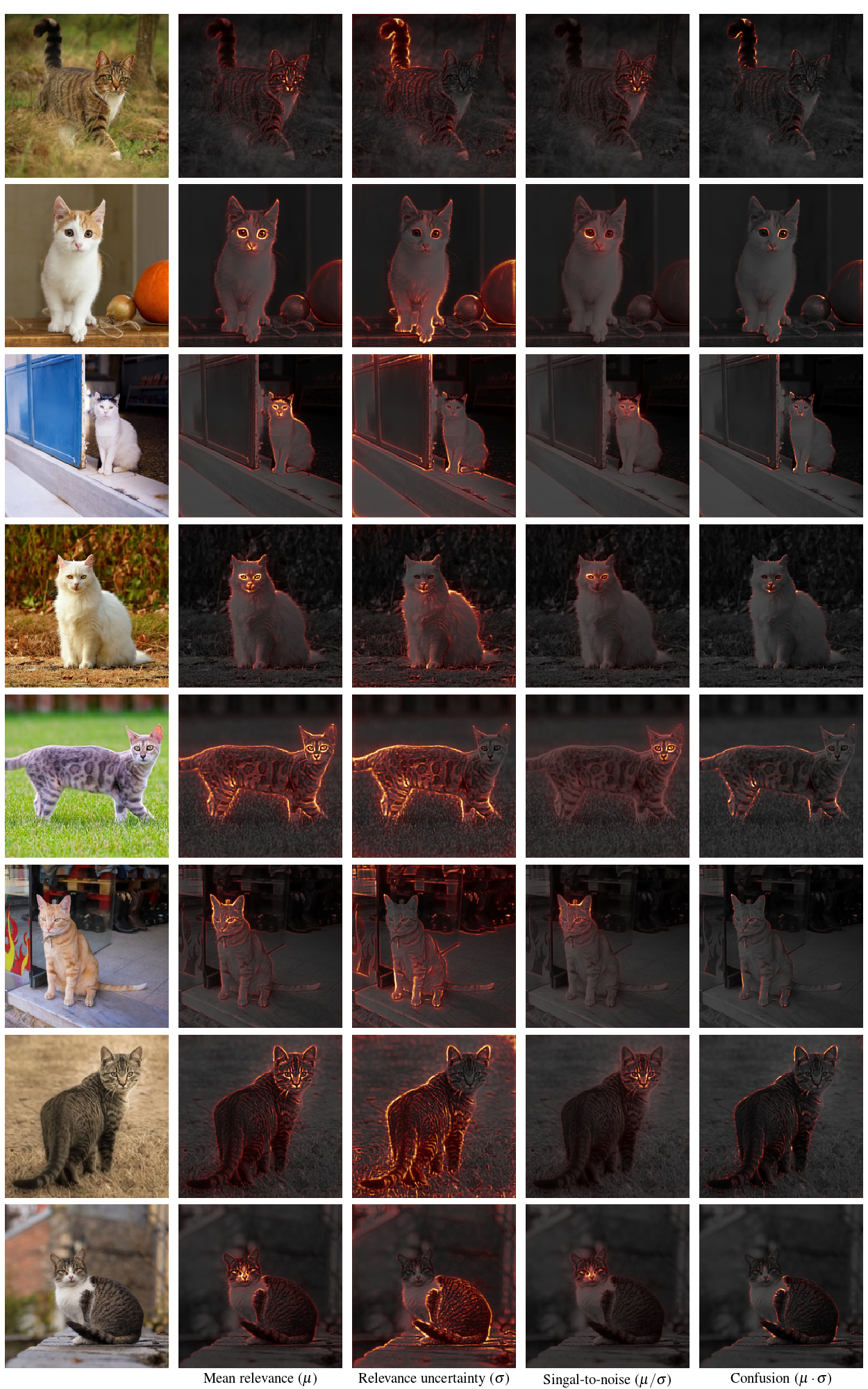}
\end{figure}

\begin{figure}[!htb]
	\centering
    \includegraphics[width=\imagesize]{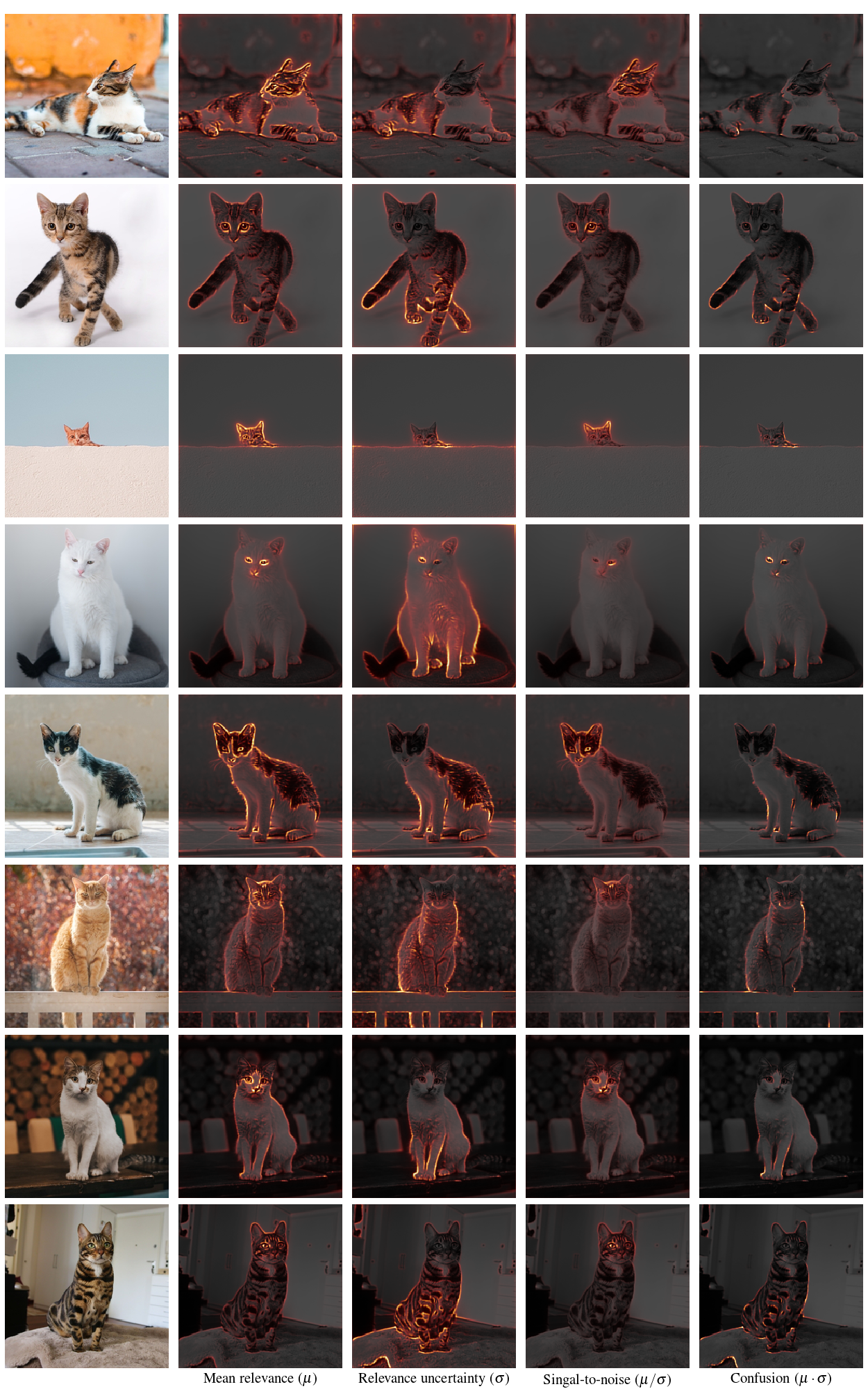}
\end{figure}

\begin{figure}[!htb]
	\centering
    \includegraphics[width=\imagesize]{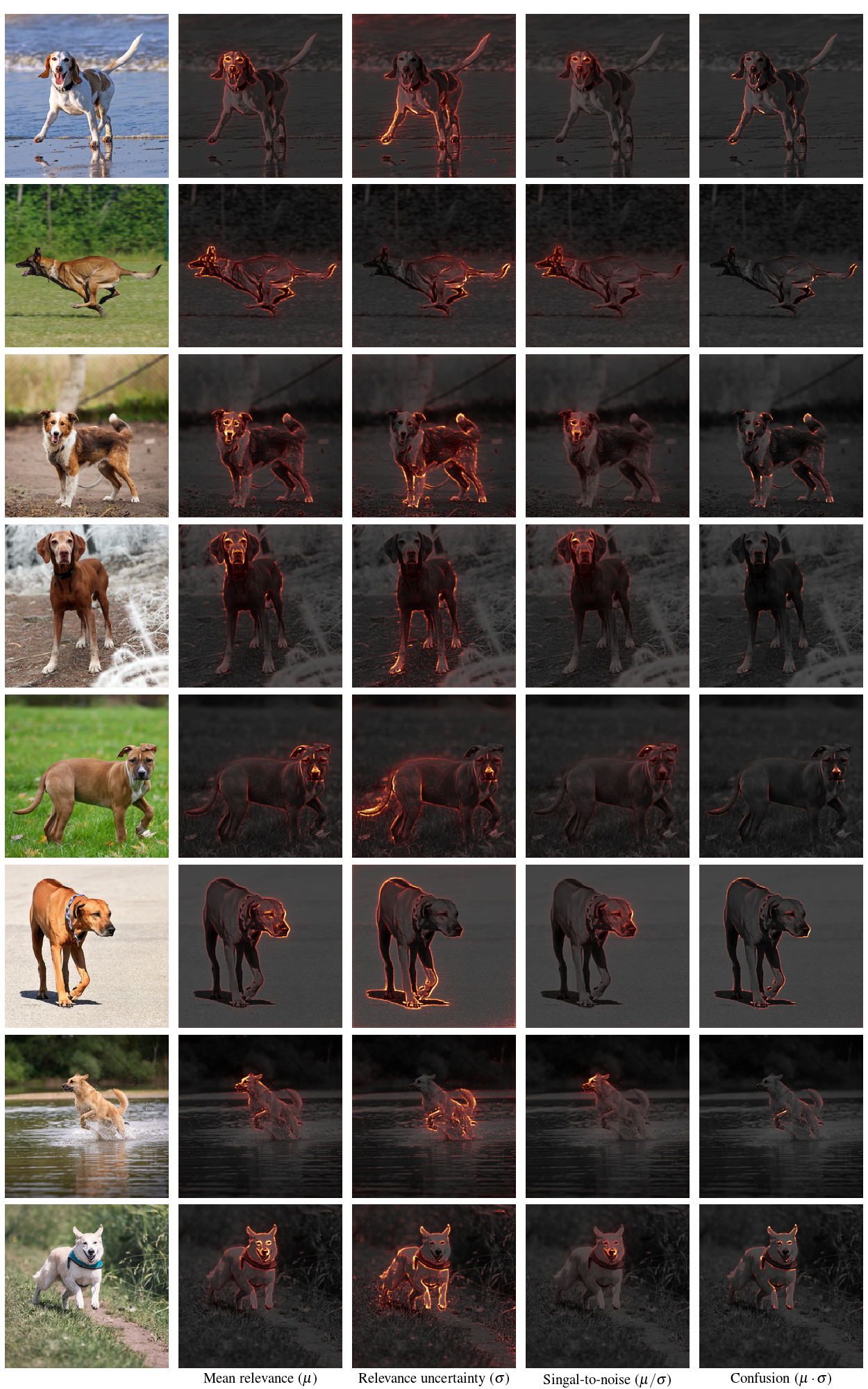}
\end{figure}

\begin{figure}[!htb]
	\centering
    \includegraphics[width=\imagesize]{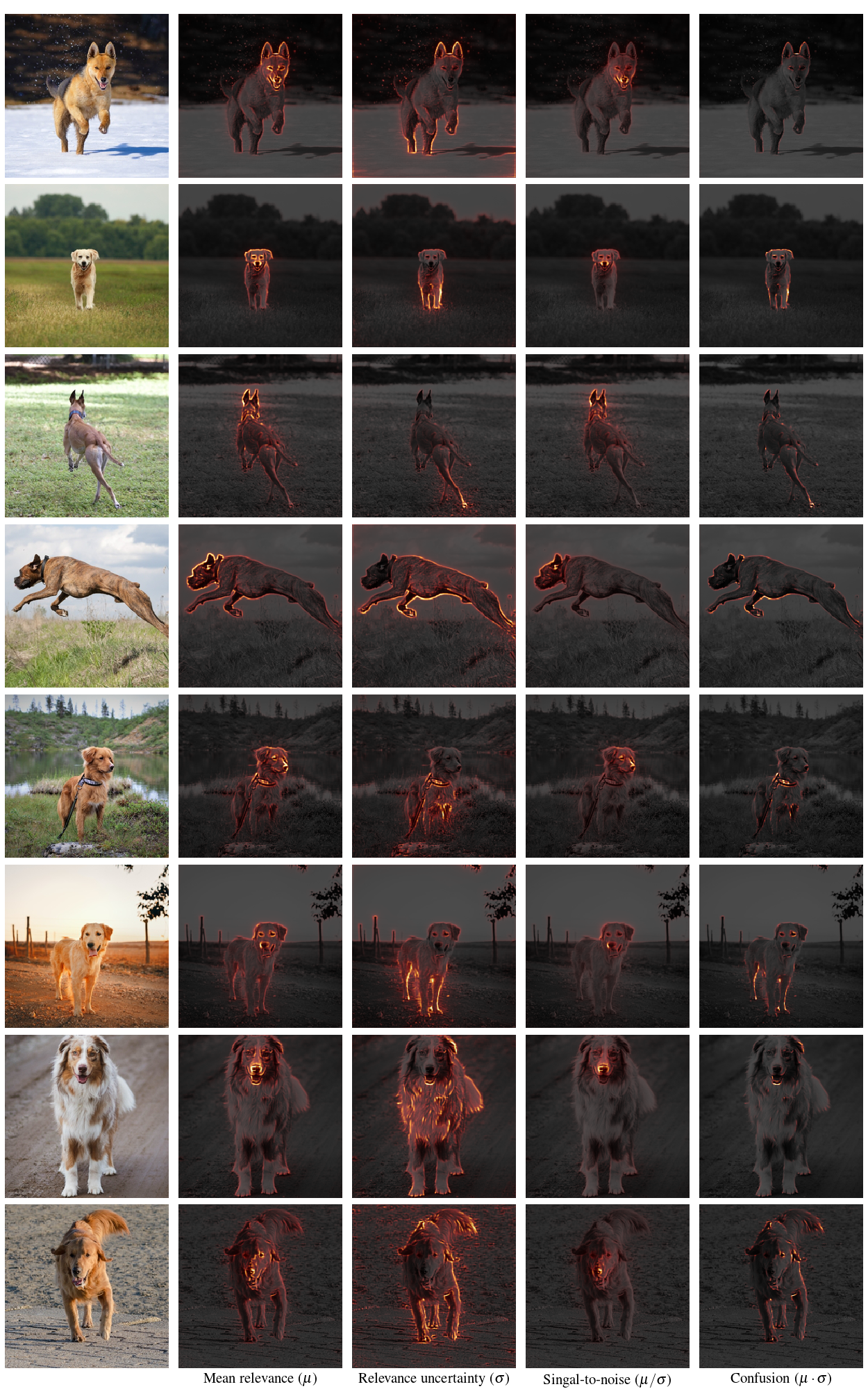}
\end{figure}

\begin{figure}[!htb]
	\centering
    \includegraphics[width=\imagesize]{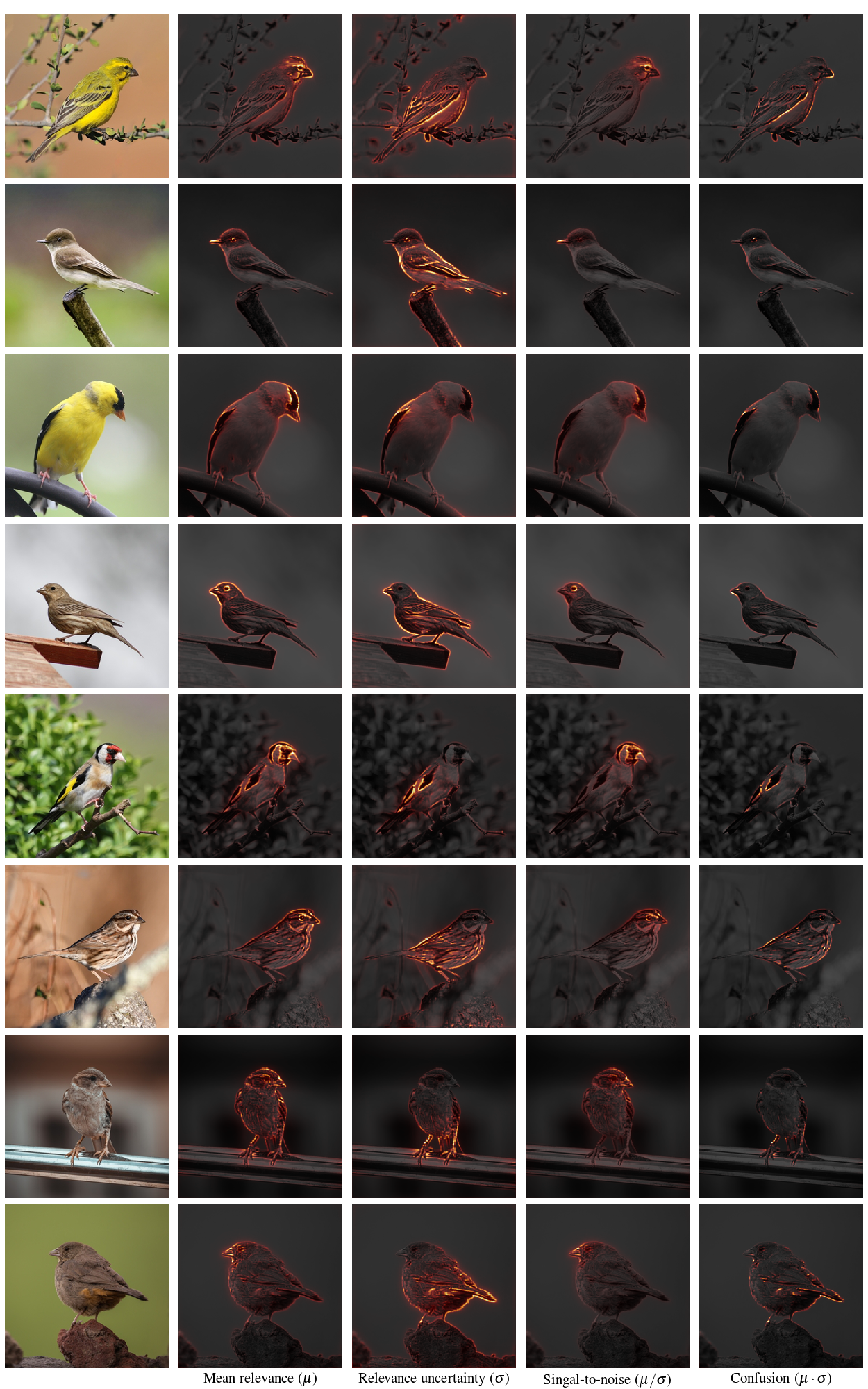}
\end{figure}

\begin{figure}[!htb]
	\centering
    \includegraphics[width=\imagesize]{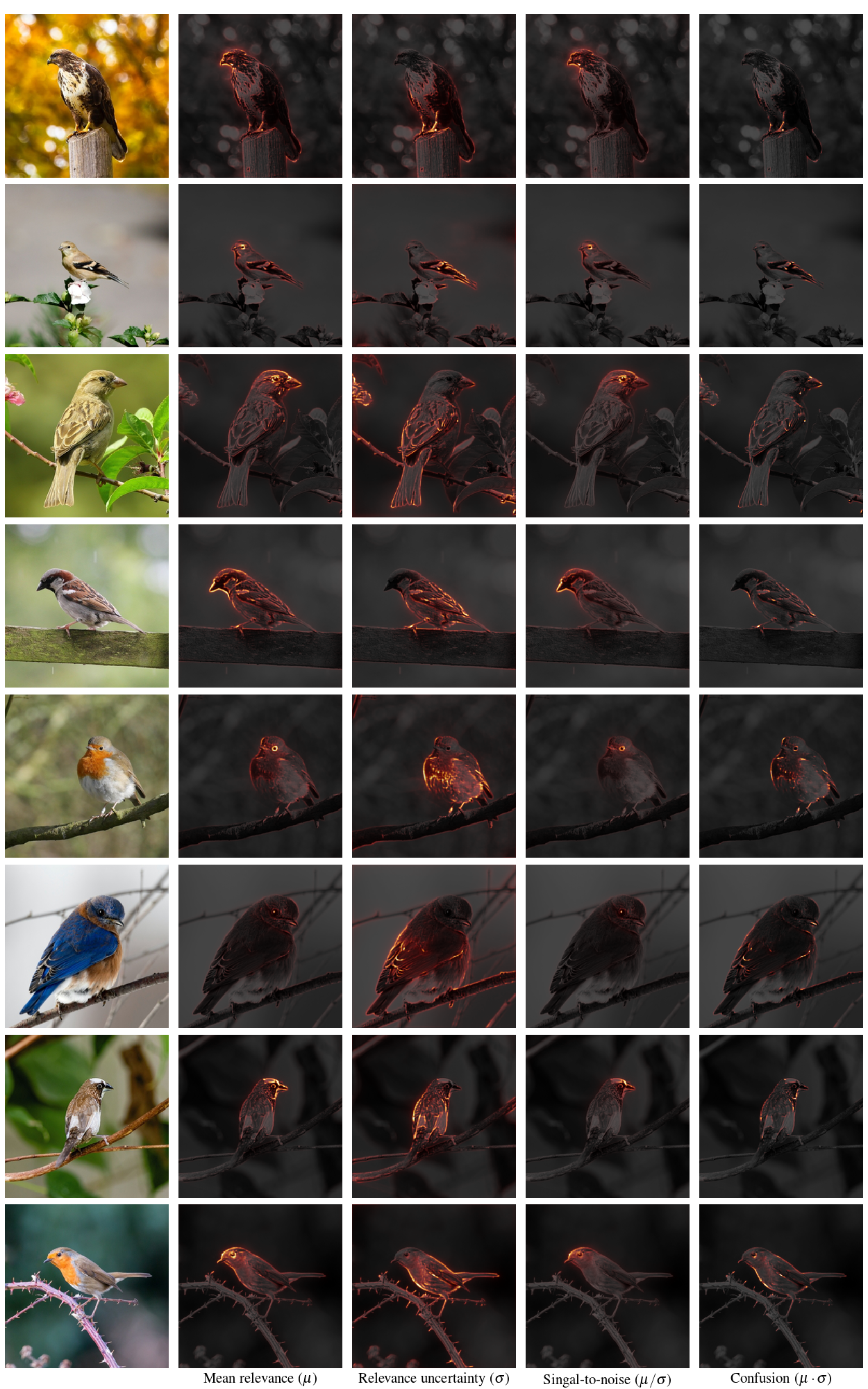}
\end{figure}

\begin{figure}[!htb]
	\centering
    \includegraphics[width=\imagesize]{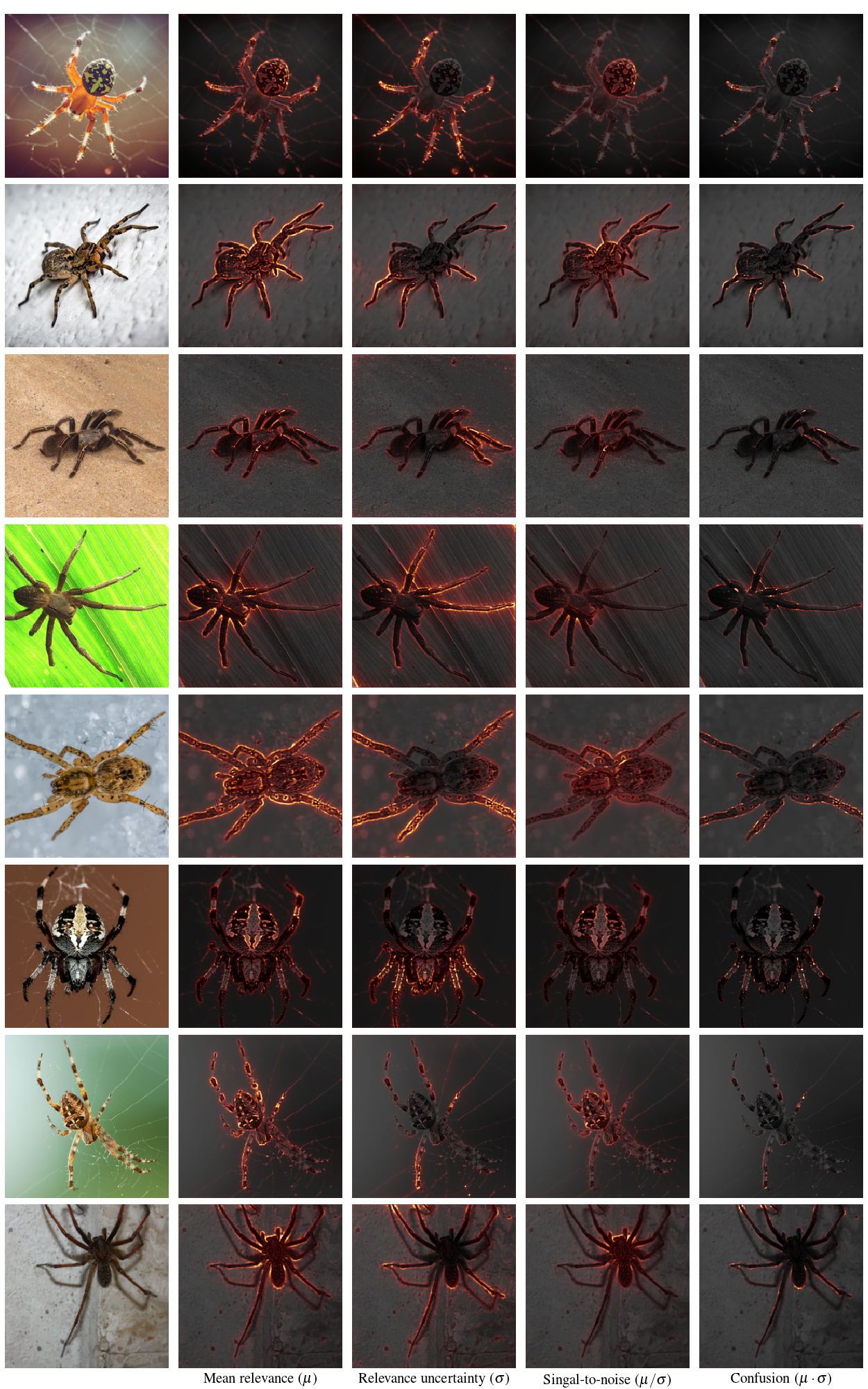}
\end{figure}

\begin{figure}[!htb]
	\centering
    \includegraphics[width=\imagesize]{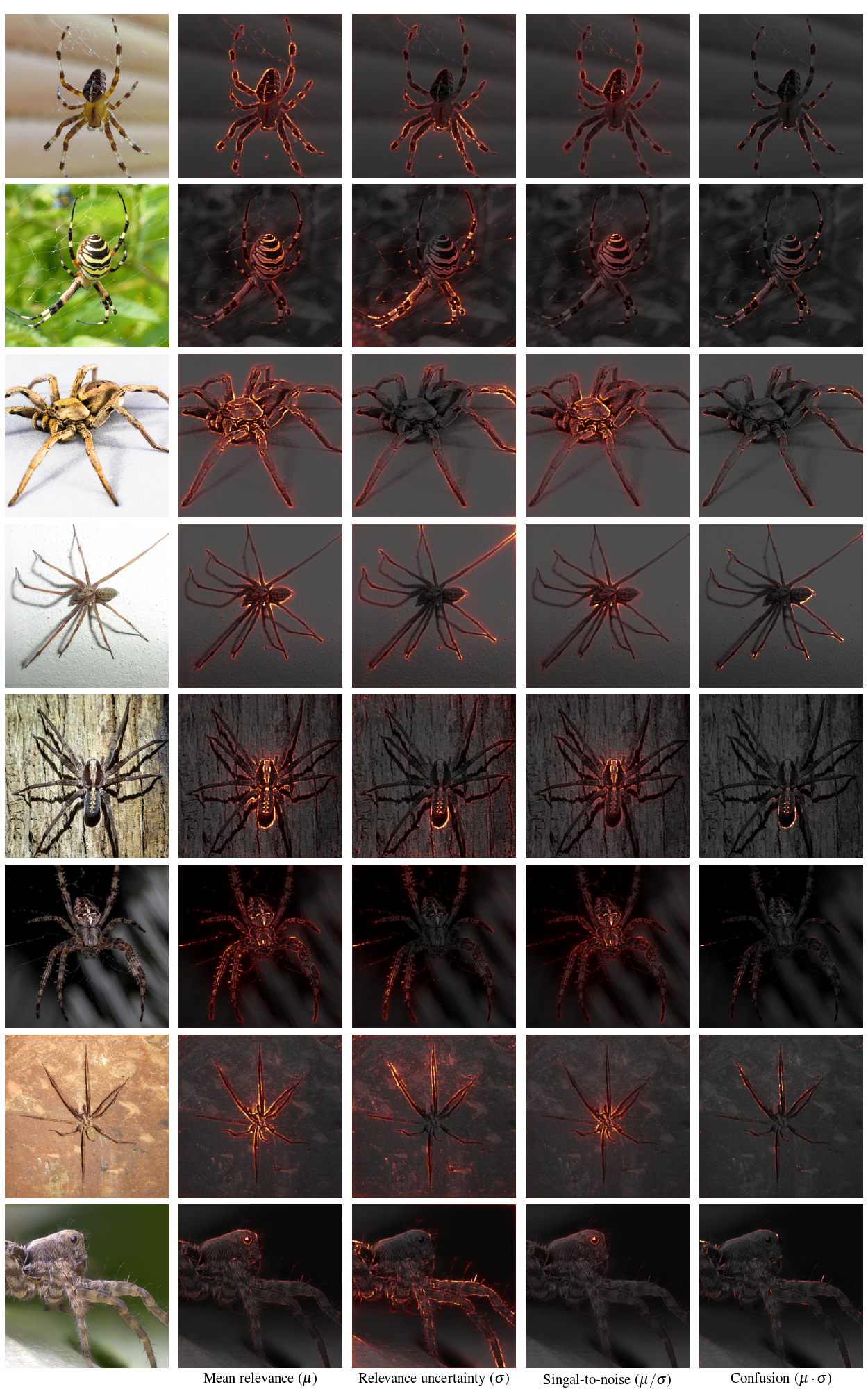}
\end{figure}
\end{document}